\documentclass{article}

\usepackage{arxiv}

\usepackage[utf8]{inputenc} 
\usepackage[T1]{fontenc}    
\usepackage{hyperref}       
\usepackage{url}            
\usepackage{booktabs}       
\usepackage{amsfonts}       
\usepackage{nicefrac}       
\usepackage{microtype}      
\usepackage{graphicx}
\usepackage{float}
\usepackage{threeparttable}
\usepackage{tabularx}
\usepackage{adjustbox}
\usepackage{subcaption}
\usepackage{natbib}

\usepackage{doi}

\title{Imagery Dataset for Condition Monitoring of Synthetic Fibre Ropes}

\author{ \href{https://orcid.org/0000-0000-0000-0000}{\includegraphics[scale=0.06]{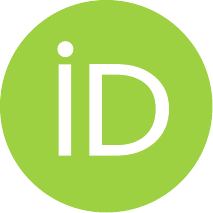}\hspace{1mm}Anju Rani}\thanks{Use footnote for providing further
		information about author (webpage, alternative
		address)---\emph{not} for acknowledging funding agencies.} \\
	Department of Energy Technology\\
	Aalborg University\\
	Esbjerg, Denmark 6700 \\
	\texttt{aran@energy.aau.dk} \\
	\And
	\href{https://orcid.org/0000-0000-0000-0000}{\includegraphics[scale=0.06]{orcid.pdf}\hspace{1mm}Daniel O.~Arroyo} \\
	Department of Energy Technology\\
	Aalborg University\\
	Esbjerg, Denmark 6700 \\
	\texttt{doa@energy.aau.dk} \\
	\And
	\href{https://orcid.org/0000-0000-0000-0000}{\includegraphics[scale=0.06]{orcid.pdf}\hspace{1mm}Petar Durdevic} \\
	Department of Energy Technology\\
	Aalborg University\\
	Esbjerg, Denmark 6700 \\
	\texttt{pdl@energy.aau.dk} \\
}

\renewcommand{\shorttitle}{Imagery Dataset for Condition Monitoring of Synthetic Fibre Ropes}

\hypersetup{
pdftitle={A template for the arxiv style},
pdfsubject={q-bio.NC, q-bio.QM},
pdfauthor={David S.~Hippocampus, Elias D.~Striatum},
pdfkeywords={First keyword, Second keyword, More},
}

\begin{document}
\maketitle

\begin{abstract}
Automatic visual inspection of synthetic fibre ropes (SFRs) is a challenging task in the field of offshore, wind turbine industries, etc. The presence of any defect in SFRs can compromise their structural integrity and pose significant safety risks. Due to the large size and weight of these ropes, it is often impractical to detach and inspect them frequently. Therefore, there is a critical need to develop efficient defect detection methods to assess their remaining useful life (RUL). To address this challenge, a comprehensive dataset has been generated, comprising a total of 6,942 raw images representing both normal and defective SFRs. The dataset encompasses a wide array of defect scenarios which may occur throughout their operational lifespan, including but not limited to placking defects, cut strands, chafings, compressions, core outs and normal. This dataset serves as a resource to support computer vision applications, including object detection, classification, and segmentation, aimed at detecting and analyzing defects in SFRs. The availability of this dataset will facilitate the development and evaluation of robust defect detection algorithms. The aim of generating this dataset is to assist in the development of automated defect detection systems that outperform traditional visual inspection methods, thereby paving the way for safer and more efficient utilization of SFRs across a wide range of applications.
\end{abstract}


\section*{Specifications Table} 
\begin{tabular}{p{4cm}p{11cm}}  
\hline
Subject                        & Computer Vision and Pattern Recognition \\
Specific subject area          & Computer vision methods (CVM) for detection of anomalies in synthetic fibre ropes (SFRs)  \\
Type of data                   & Image (png)  \\
How data were acquired         & Data was collected using a Basler acA2000 camera with a Basler C11-5020-12M-P Premium 12-megapixel lens. The experimental setup consists of a motor, three sheaves (one sheave for holding weight, two rotation pulley blocks, two wire guide wheels), four Aputure AL-MC RGBWW LED lights, NVIDIA Jetson Nano P3450, and ten SFRs each of length 8 cm subjected to a weight of 50 kg. To ensure a comprehensive dataset, images were captured at different motor speeds, enabling the capture of a maximum number of defects. This variation in motor speed helps simulate different operating conditions that the SFRs may experience. The dataset images were captured at a frame rate of 165 frames per second (FPS) and had a resolution of 2000 x 1080 pixels.  \\
Data format                    & Raw   \\
Description of data collection & Ten different SFRs each having a length of 8 m have been used for collecting datasets. Nine ropes used for the 
                                 experiment contain multiple defects of each class ranging between high, medium and low while one rope is a non-defective rope. The dataset includes various classes such as normal, compression, core out, placking (high, low, medium), chafing (high, low, medium), and cut strands (high, low, medium). These classes encompass a wide range of defects commonly found in SFRs. The primary objective of collecting the dataset is to predict and analyze the defects in SFRs to provide enhanced efficiency and reliability in various industries. \\
Data source location           & \begin{tabular}[c]{@{}l@{}}
                                Institution: Department of Energy, Aalborg University\\ 
                                City/Region: Esbjerg\\ 
                                Country: Denmark \\
                                \end{tabular}  \\
Data accessibility             & \begin{tabular}[c]{@{}l@{}}
                                Repository name: Mendeley Data \\ 
                                Data identification number: \url{10.17632/by9wy6fxsr} (Version 1, Version 2)\\ 
                                Direct URL to data: \url{https://data.mendeley.com/datasets/by9wy6fxsr} \\ 
                                \end{tabular} \\ 
\hline
\end{tabular}

\begin{table}[h]
\caption{Description of defects and defect class in SFRs dataset}
\centering
\label{table:Defect_class}
 \begin{adjustbox}{width=\textwidth}
\begin{tabular}{cccc}
\hline
\textbf{Damage Type} & \textbf{Description} & \textbf{Damage Level}             & \textbf{Count}  \\ \hline
                     
                     &                       & High                             & 170 \\
Placking             & \multicolumn{1}{m{8.3cm}}{Loop of fibre raised on the surface of rope by being snagged on a 
                                                pointed object}              
                                             & Medium                           & 167 \\
                     &                       & Low                              & 200 \\ 
                                                         
                     &                       & High                             &  253 \\
Cut Strands          & \multicolumn{1}{m{8.3cm}}{Damage caused due to incisions with a sharp edge on the surface of 
                                               the fibre rope}
                                             & Medium                           & 212  \\
                     &                       & Low                              & 226  \\ 
                     
                     &                        & High                            & 201  \\
Chafing              & \multicolumn{1}{m{8.3cm}}{Planing of the rope surface as the result of scraping with a sharp 
                                                 or abrasive edge}    
                                              & Medium                          & 195  \\
                     &                        & Low                             & 252  \\ 
                     
                     &                        & Only compression                & 1436 \\   
Compression          & \multicolumn{1}{m{8.3cm}}{Change in diameter in comparison to normal rope indicating core or 
                                                 internal damage caused due to overloading or shock loads}
                                              & Compression with chafing        & 238  \\
                     &                        & Compression with cut strands    & 254  \\

Core out            & \multicolumn{1}{m{8.3cm}}{Manufacturing defect where the core used in the centre of strands is visible on the surface of the 
                                                fibre rope}  
                                              & Only core out                   & 739  \\ 
                    &                         & Core out with cut strands       & 49   \\
                                              
Normal              & \multicolumn{1}{m{8.3cm}}{Rope with no visible damage} & - & 2142 \\
Extra               & \multicolumn{1}{m{8.3cm}}{Rope with a marking which must not be used for generative networks due to generation of incorrect image datasets } & - & 208 \\
\hline
\end{tabular}
\end{adjustbox}
\end{table}

\section*{Value of the Data}
\begin{itemize}
\item SFRs are extensively used in ocean engineering, offshore, wind turbine industries, etc., where safety is of paramount importance. The dataset enables the development of more efficient and reliable defect detection methods, leading to improved safety and risk mitigation. Timely identification of defects can help to prevent potential failures and ensure the ropes are in optimal condition.
\item Researchers and practitioners can utilize the dataset to develop and evaluate algorithms that can accurately identify and classify various types of defects, such as plackings, cut strands, chafing, compression, core out, and normal. The dataset contains all possible defect scenarios to the best of our knowledge. Also, this is among the only publicly accessible datasets on SFRs.
\item This dataset aims to serve as a valuable resource for computer vision applications such as object detection, classification, and segmentation. Researchers can utilize the dataset to train and test algorithms for tasks related to defect detection and CM applications. This promotes the development of more accurate and efficient computer vision techniques tailored to the unique challenges of inspecting large, heavy ropes.
\item This dataset may also encourage the adoption of standardized testing protocols and best practices for the inspection of SFRs across industries.
\end{itemize}

\begin{figure}[ht]
\centering
\includegraphics[scale=0.5]{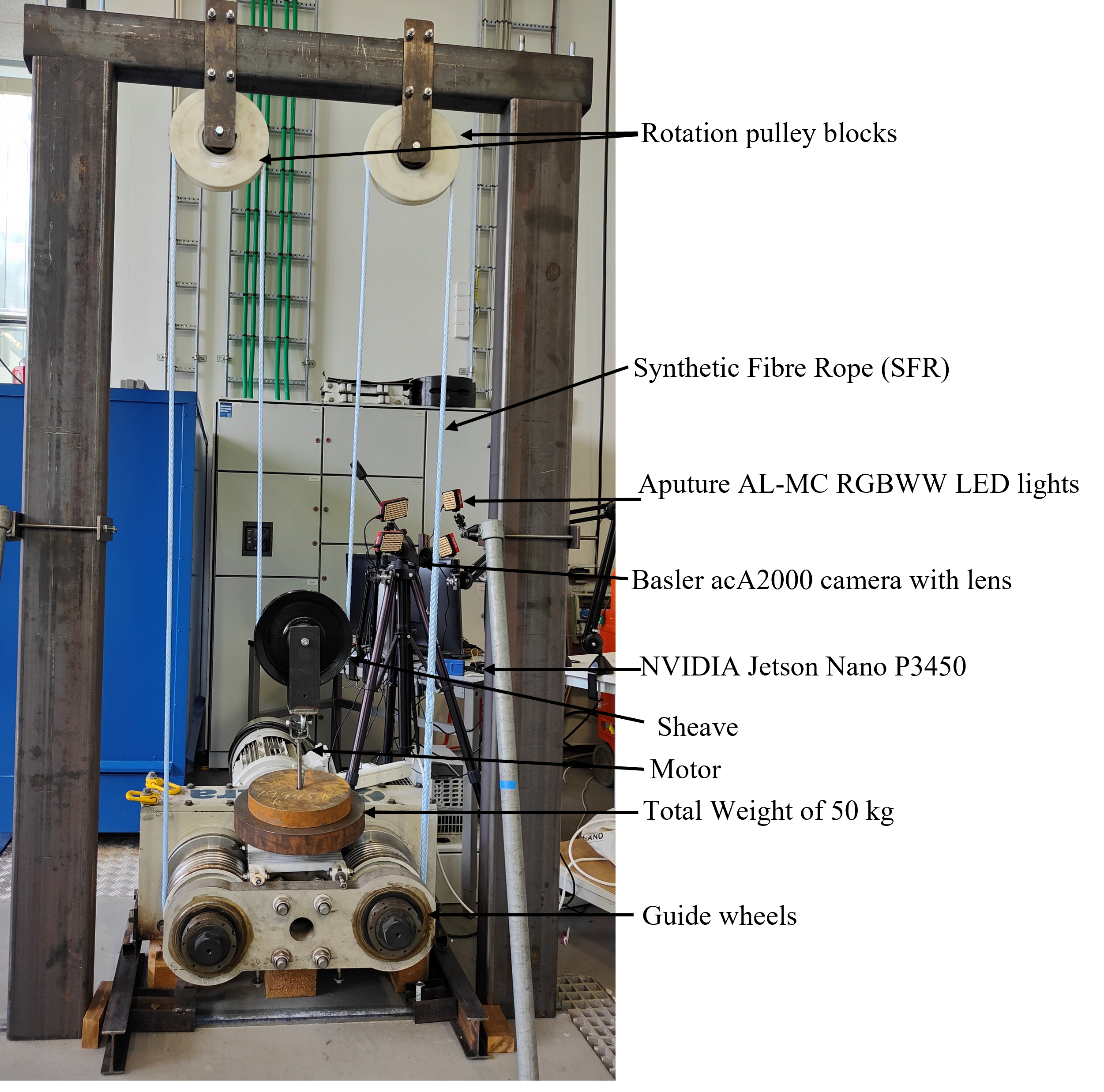}
\caption{Experimental setup for collecting SFRs dataset.}
\label{fig:Setup}
\end{figure}

\begin{figure}[ht]
     \centering
     \begin{subfigure}[b]{0.32\textwidth}
         \centering
         \includegraphics[width=\textwidth]{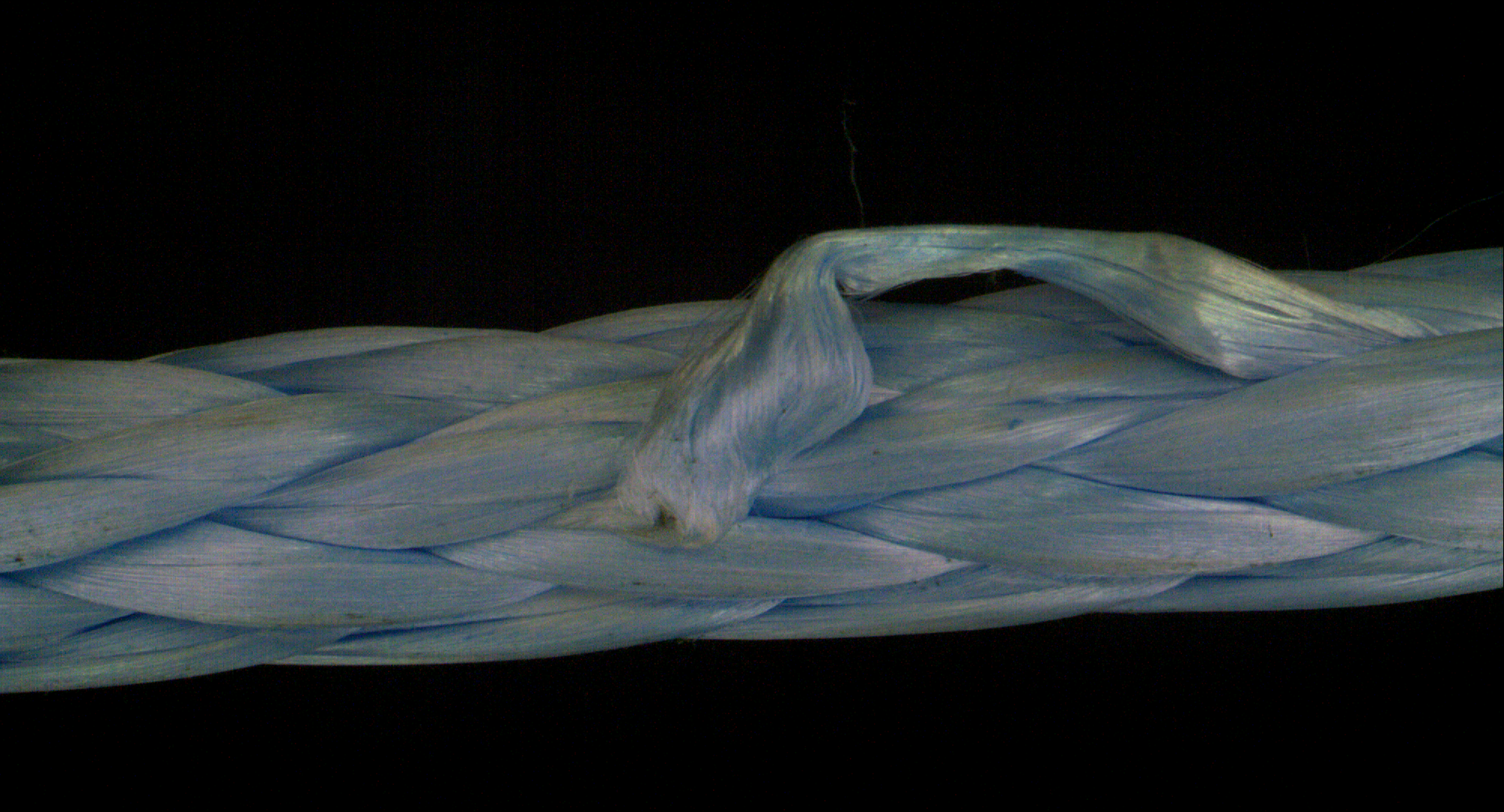}
         \caption{Placking High}
         \label{fig:Defects1}
     \end{subfigure}
     \begin{subfigure}[b]{0.32\textwidth}
         \centering
         \includegraphics[width=\textwidth]{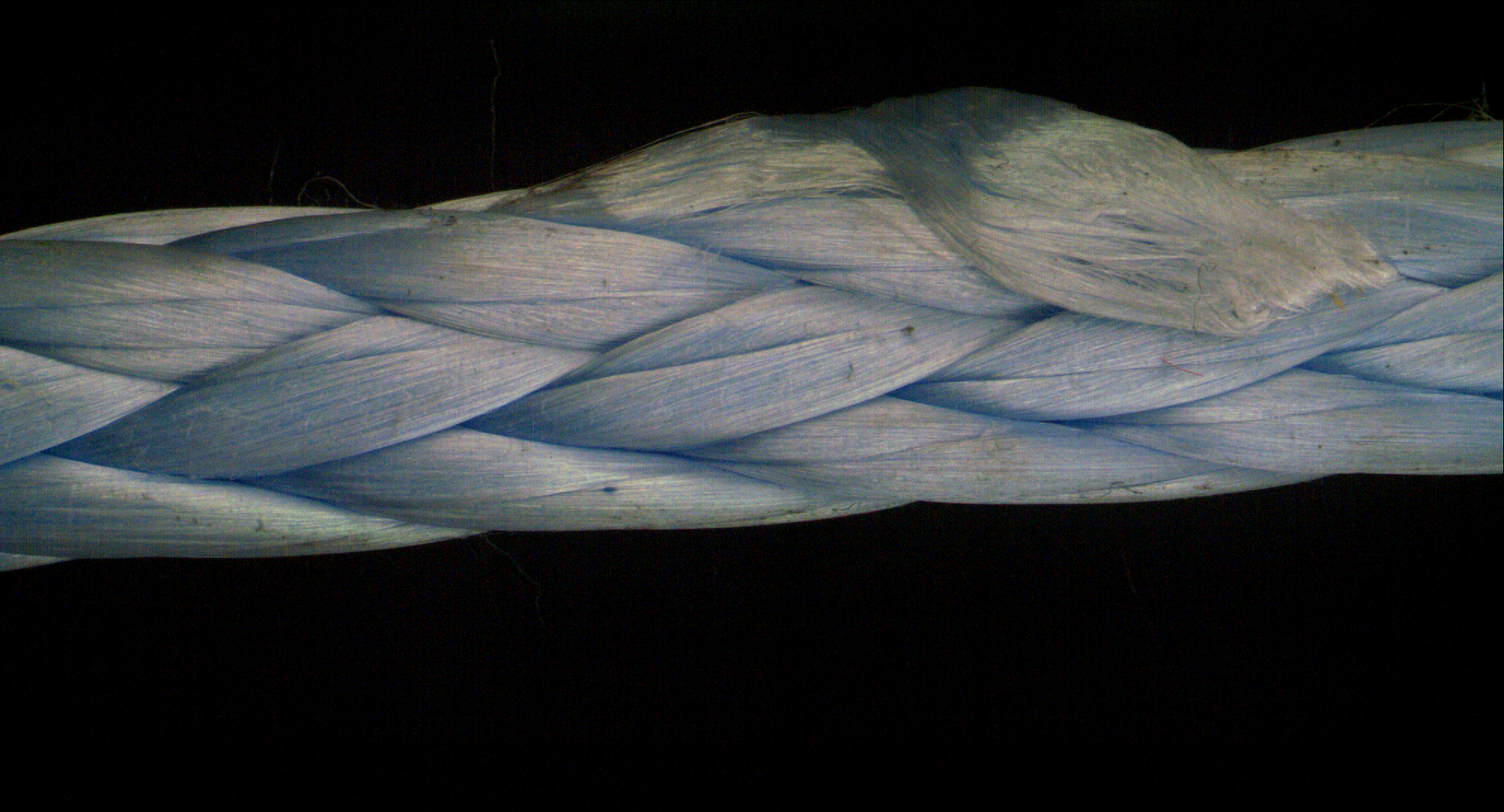}
         \caption{Placking Medium}
         \label{fig:Defects2}
     \end{subfigure}
     \begin{subfigure}[b]{0.32\textwidth}
         \centering
         \includegraphics[width=\textwidth]{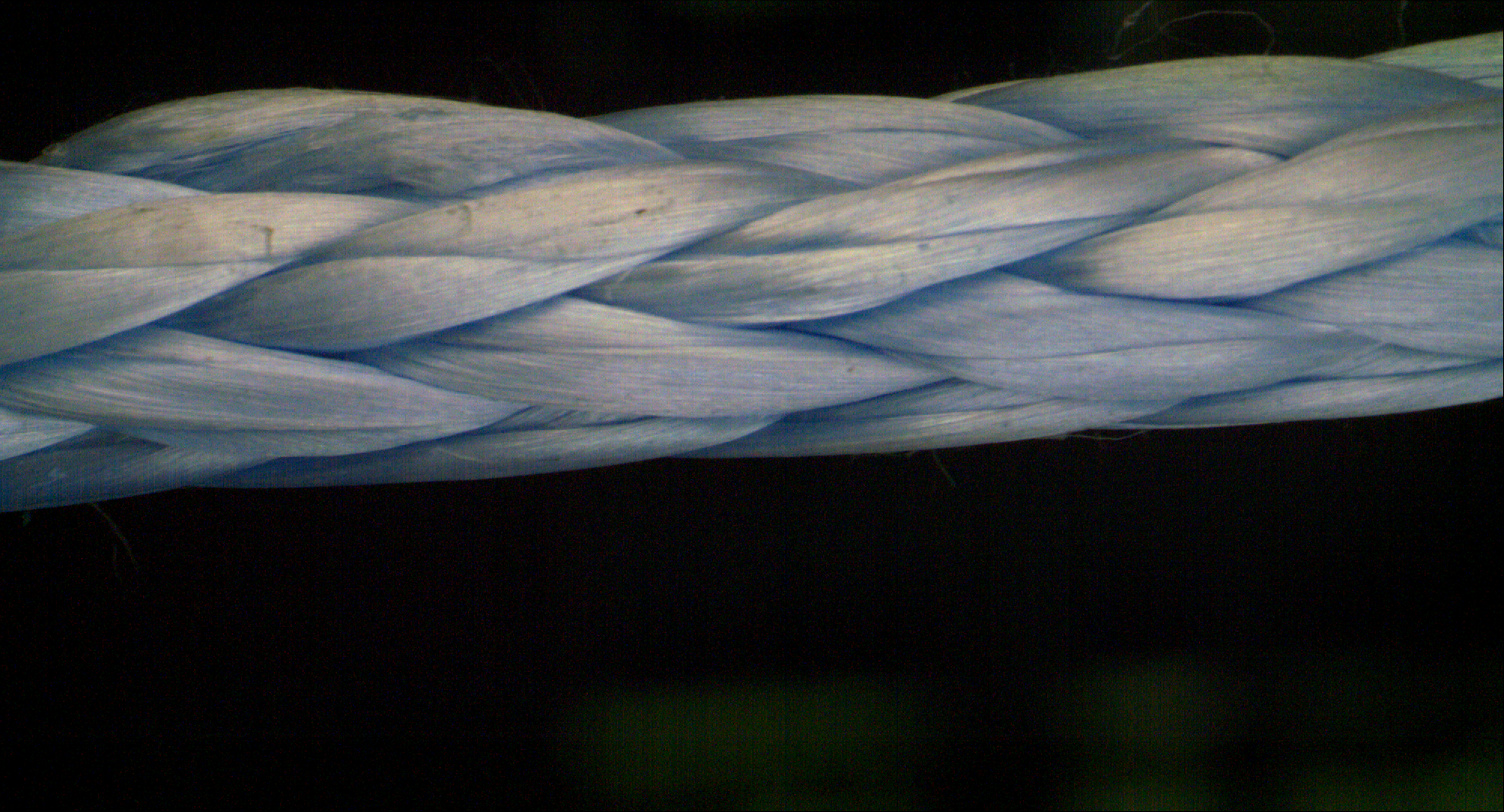}
         \caption{Placking Low}
         \label{fig:Defects3}
     \end{subfigure}
     \begin{subfigure}[b]{0.32\textwidth}
         \centering
         \includegraphics[width=\textwidth]{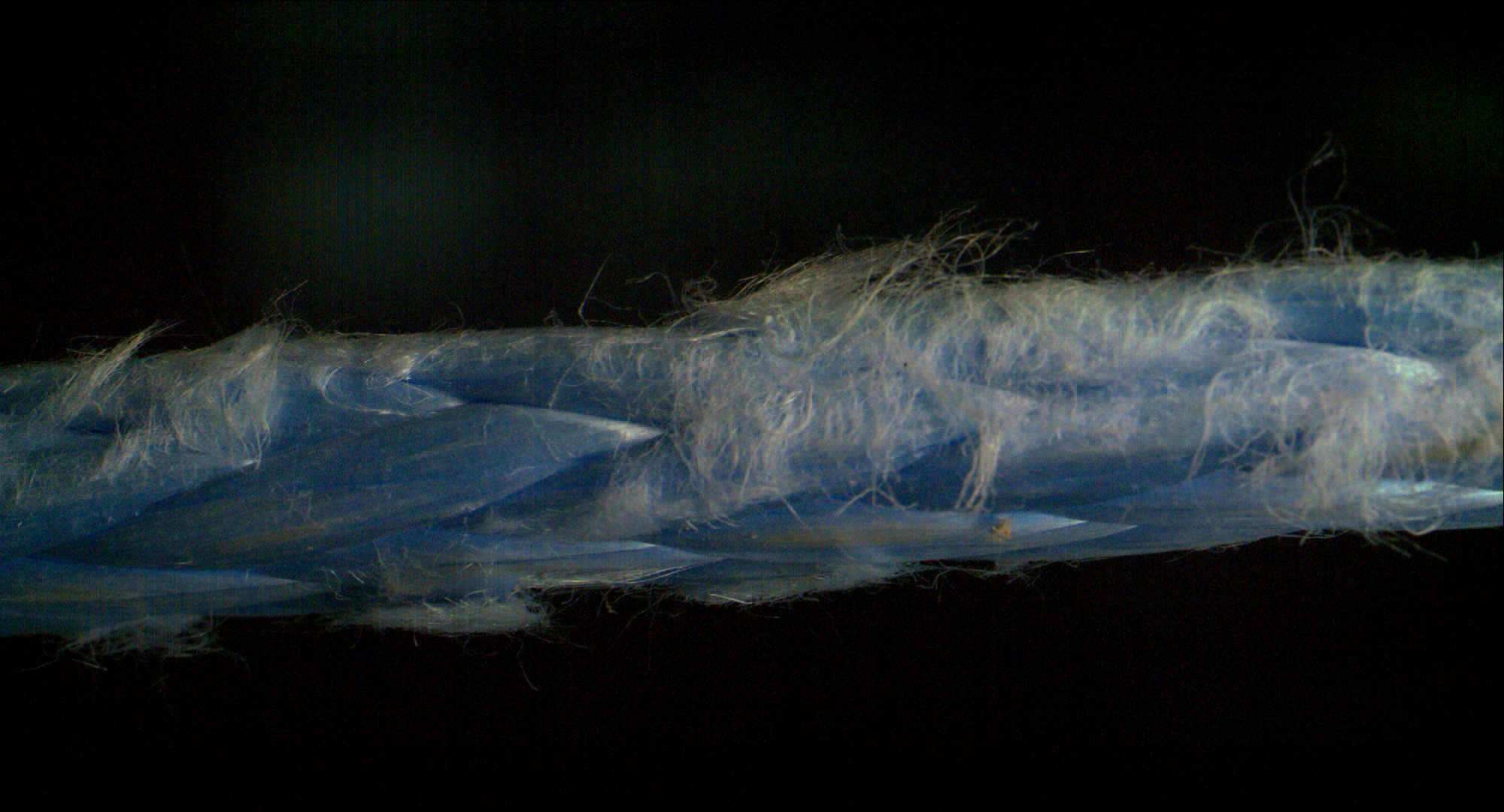}
         \caption{Chafing High}
         \label{fig:Defects4}
     \end{subfigure}
     \begin{subfigure}[b]{0.32\textwidth}
         \centering
         \includegraphics[width=\textwidth]{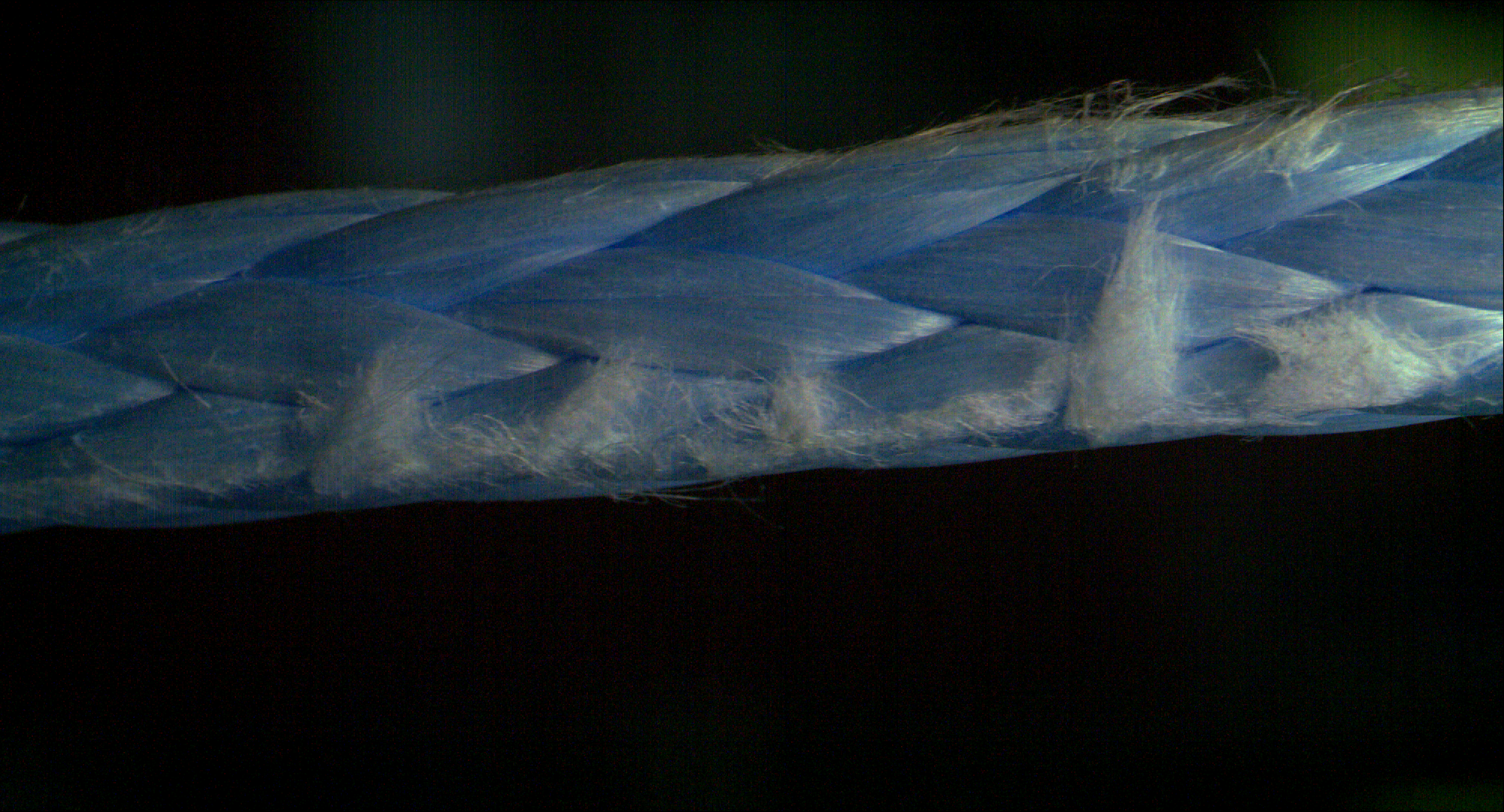}
         \caption{Chafing Medium}
         \label{fig:Defects5}
     \end{subfigure}
     \begin{subfigure}[b]{0.32\textwidth}
         \centering
         \includegraphics[width=\textwidth]{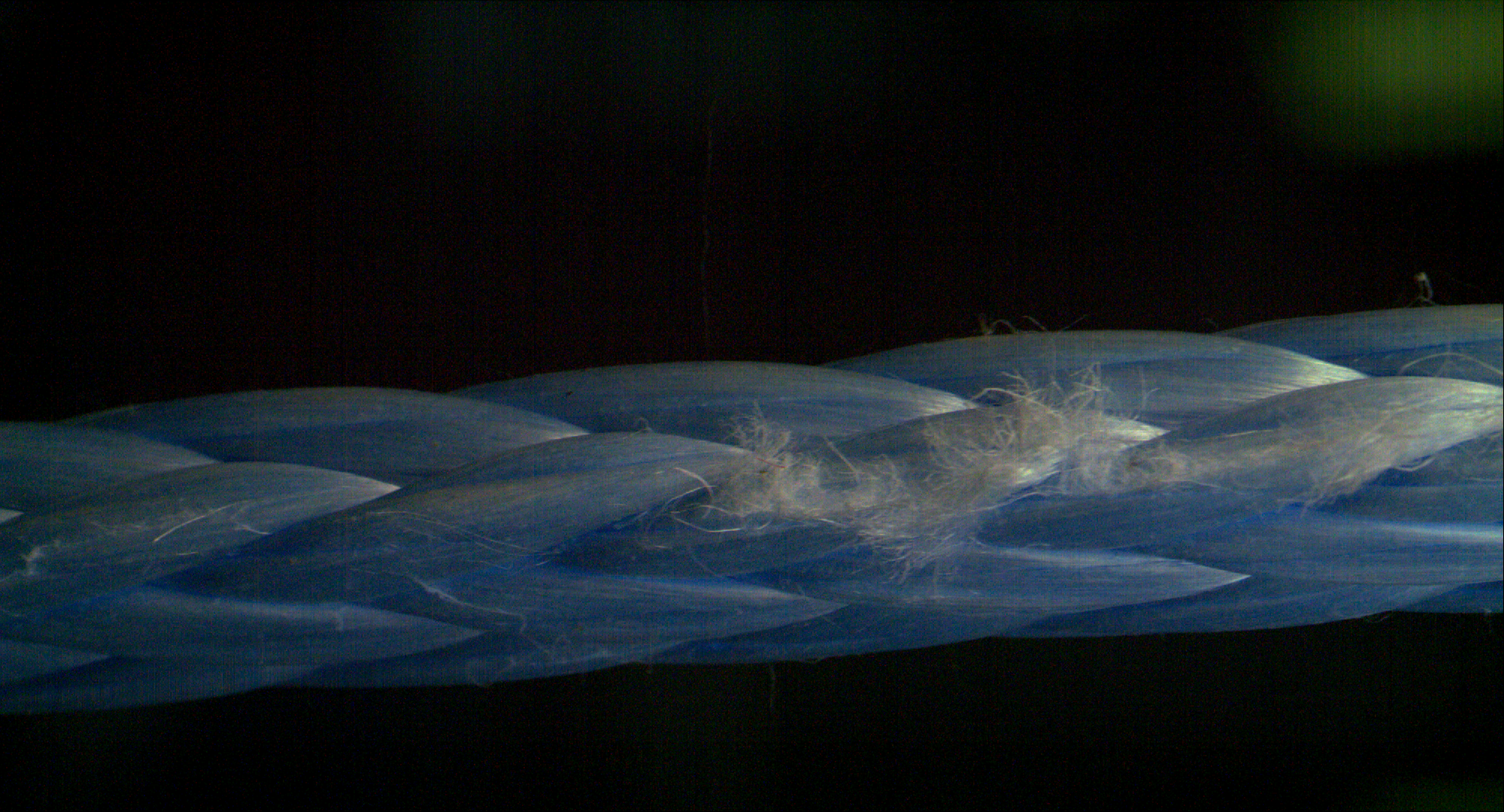}
         \caption{Chafing Low}
         \label{fig:Defects6}
     \end{subfigure}     
     \begin{subfigure}[b]{0.32\textwidth}
         \centering
         \includegraphics[width=\textwidth]{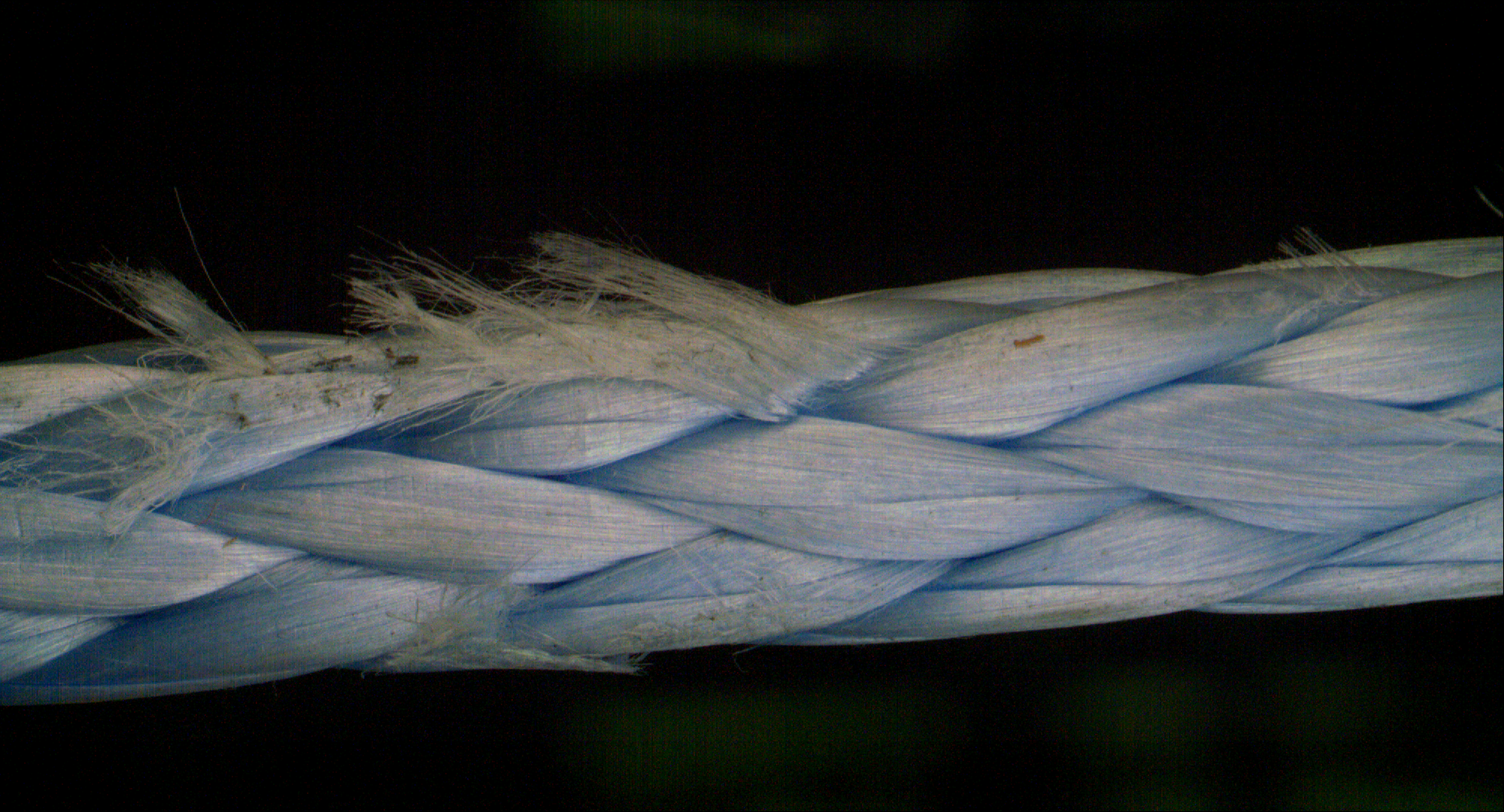}
         \caption{Cut Strands High}
         \label{fig:Defects7}
     \end{subfigure}
     \begin{subfigure}[b]{0.32\textwidth}
         \centering
         \includegraphics[width=\textwidth]{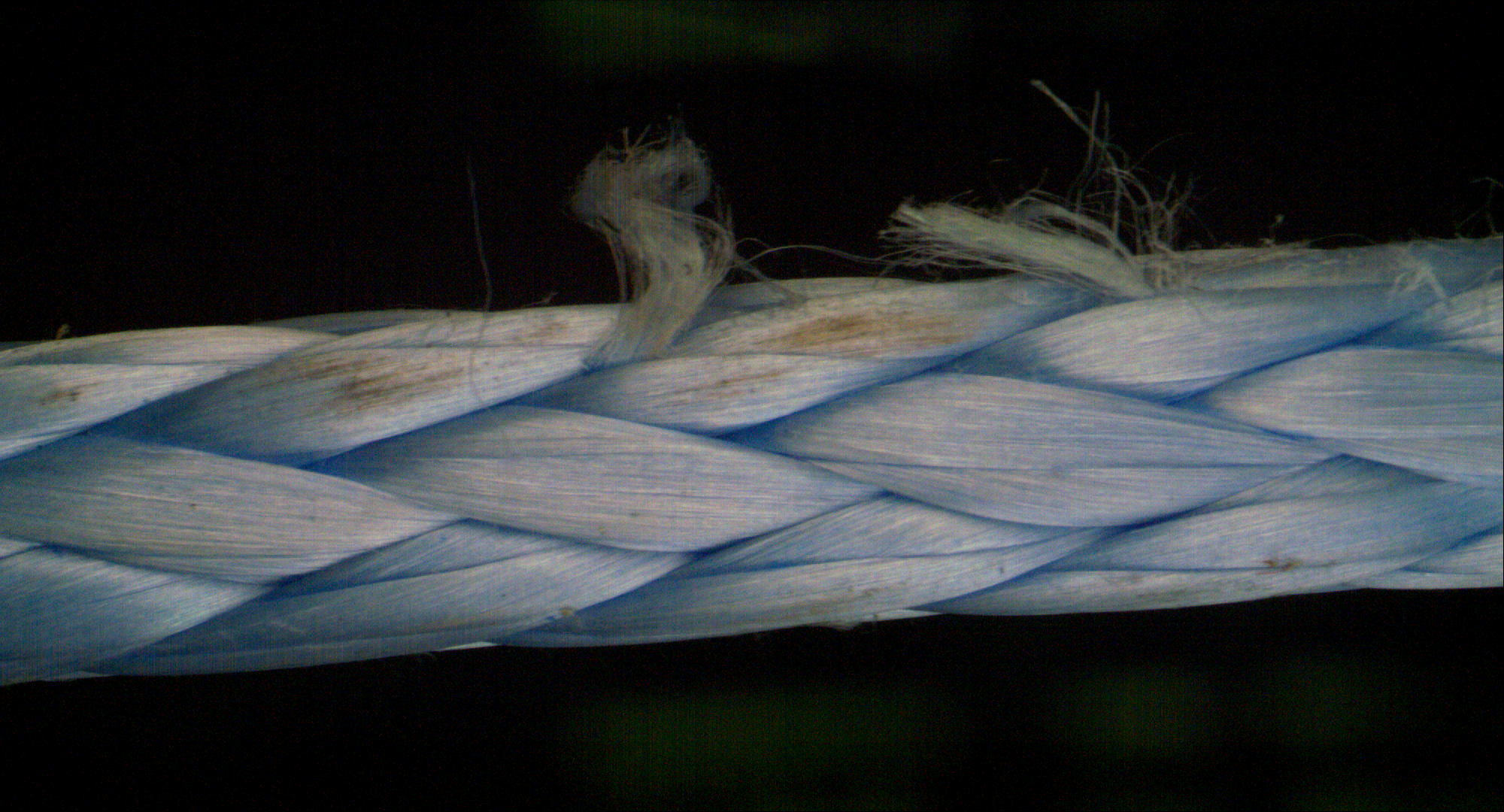}
         \caption{Cut Strands Medium}
         \label{fig:Defects8}
     \end{subfigure}
     \begin{subfigure}[b]{0.32\textwidth}
         \centering
         \includegraphics[width=\textwidth]{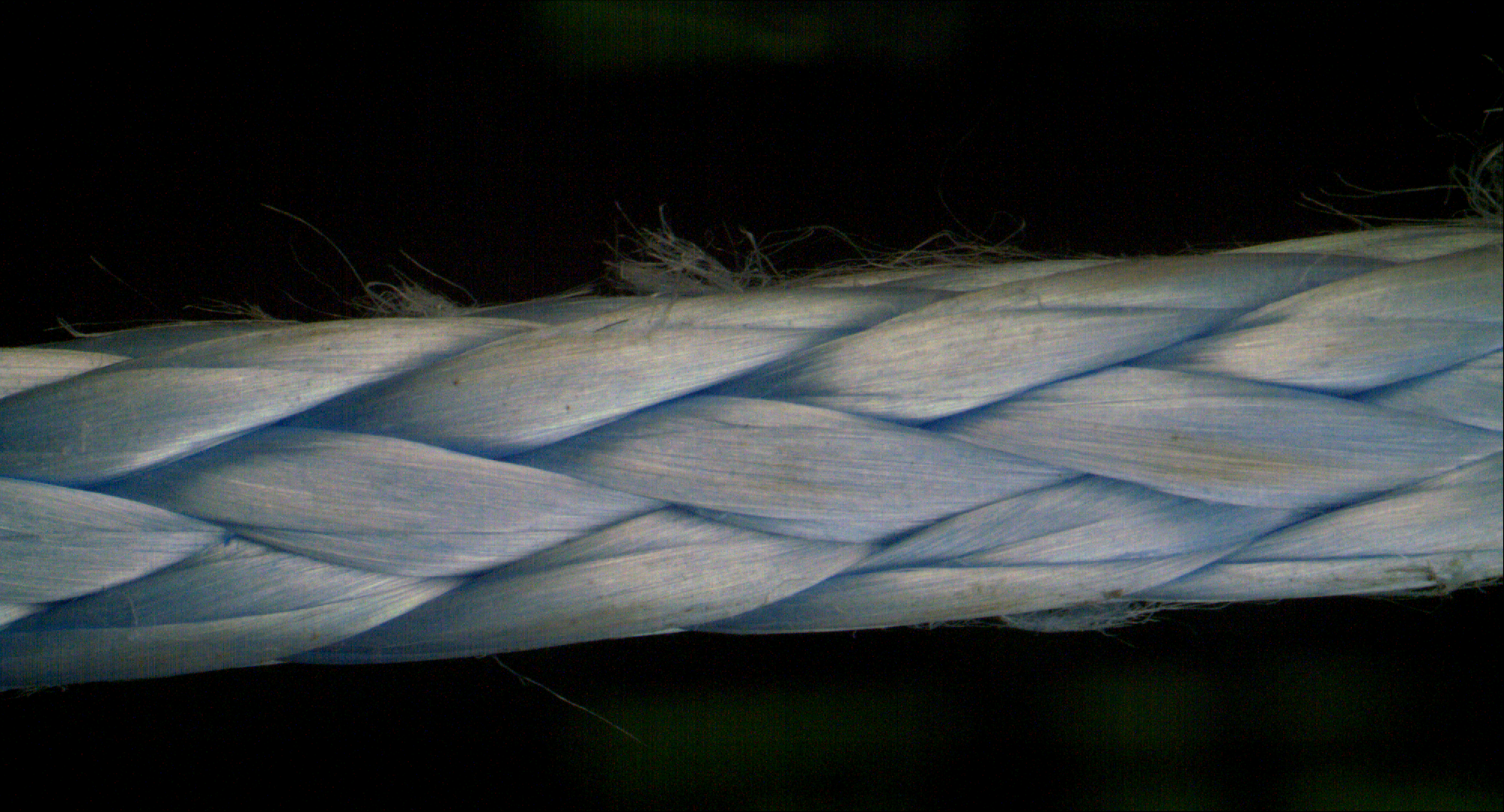}
         \caption{Cut Strands Low}
         \label{fig:Defects9}
     \end{subfigure}     
     \begin{subfigure}[b]{0.32\textwidth}
         \centering
         \includegraphics[width=\textwidth]{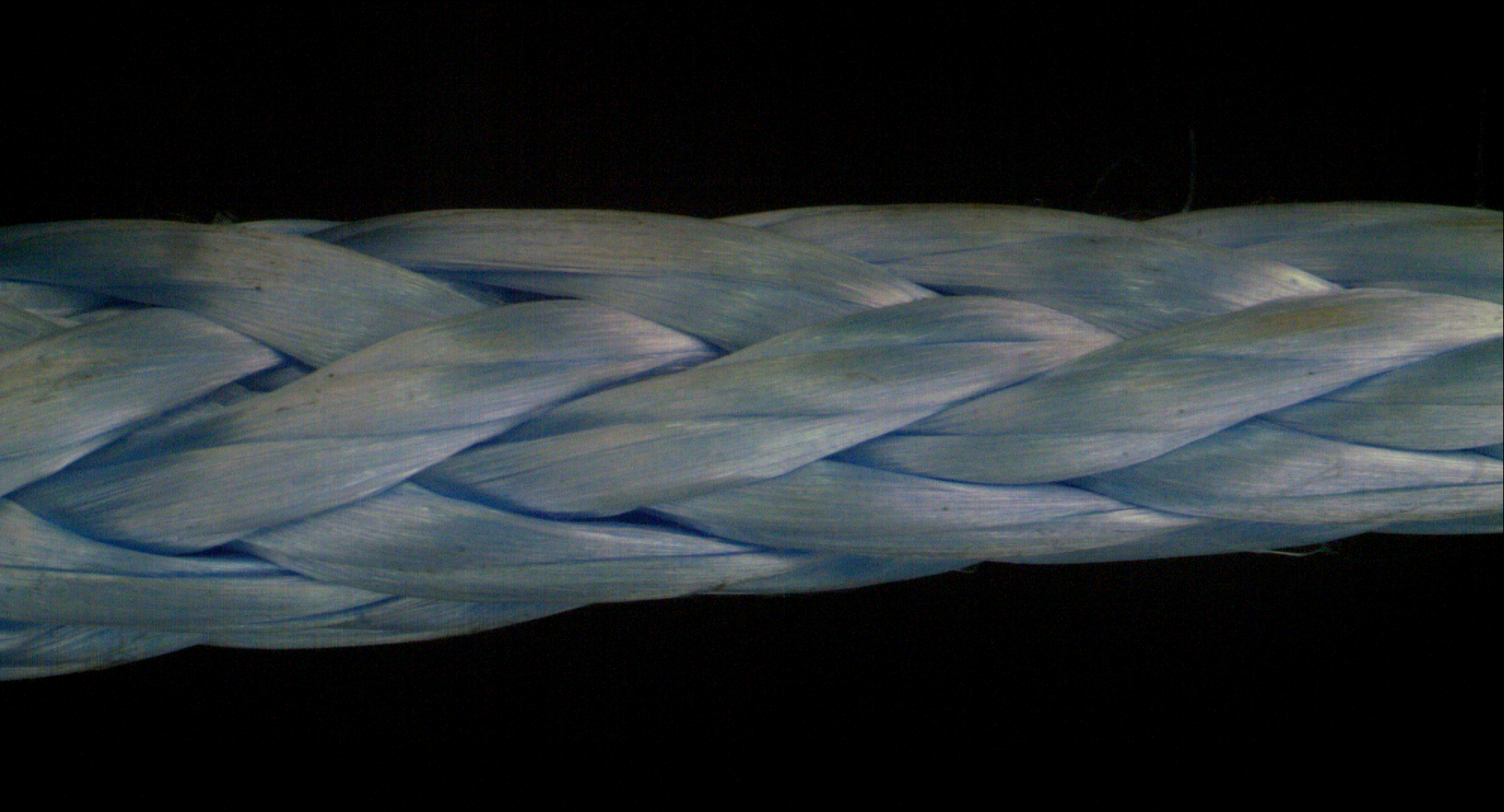}
         \caption{Compression}
         \label{fig:Defects10}
     \end{subfigure}
     \begin{subfigure}[b]{0.32\textwidth}
         \centering
         \includegraphics[width=\textwidth]{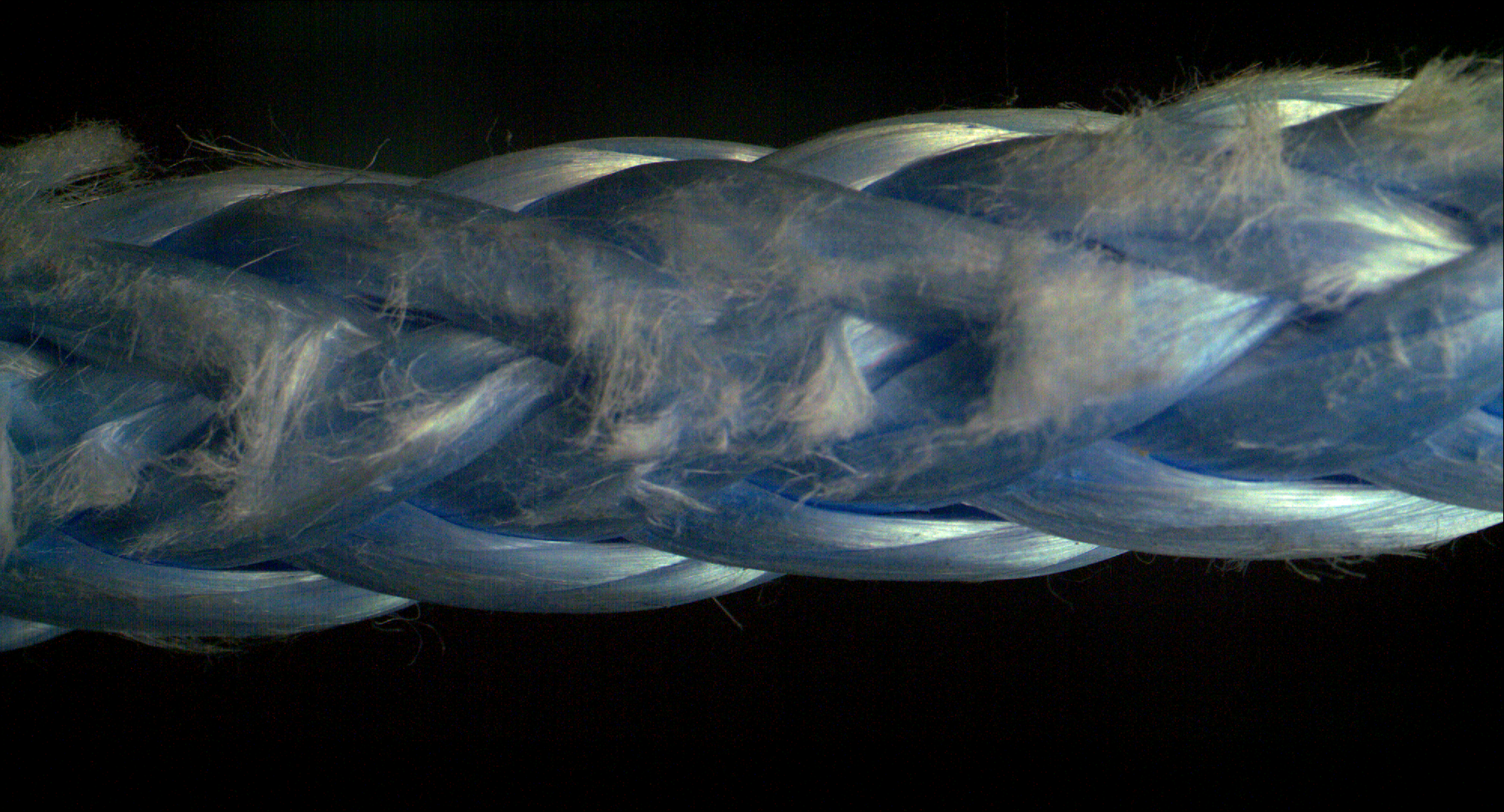}
         \caption{Compression with Chafing}
         \label{fig:Defects11}
     \end{subfigure}
          \begin{subfigure}[b]{0.32\textwidth}
         \centering
         \includegraphics[width=\textwidth]{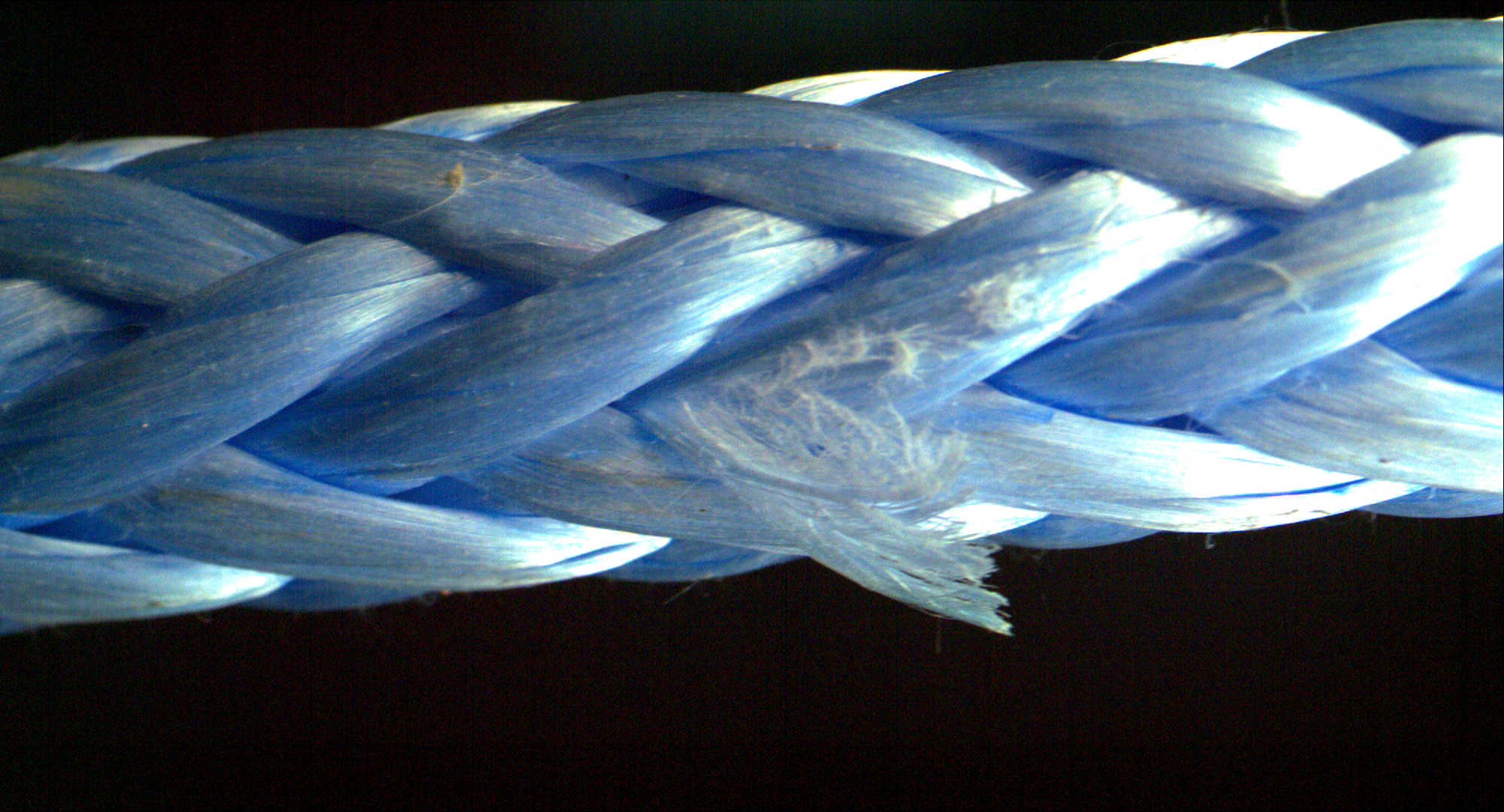}
         \caption{Compression with Cut Strands}
         \label{fig:Defects12}
     \end{subfigure}
           \begin{subfigure}[b]{0.32\textwidth}
         \centering
         \includegraphics[width=\textwidth]{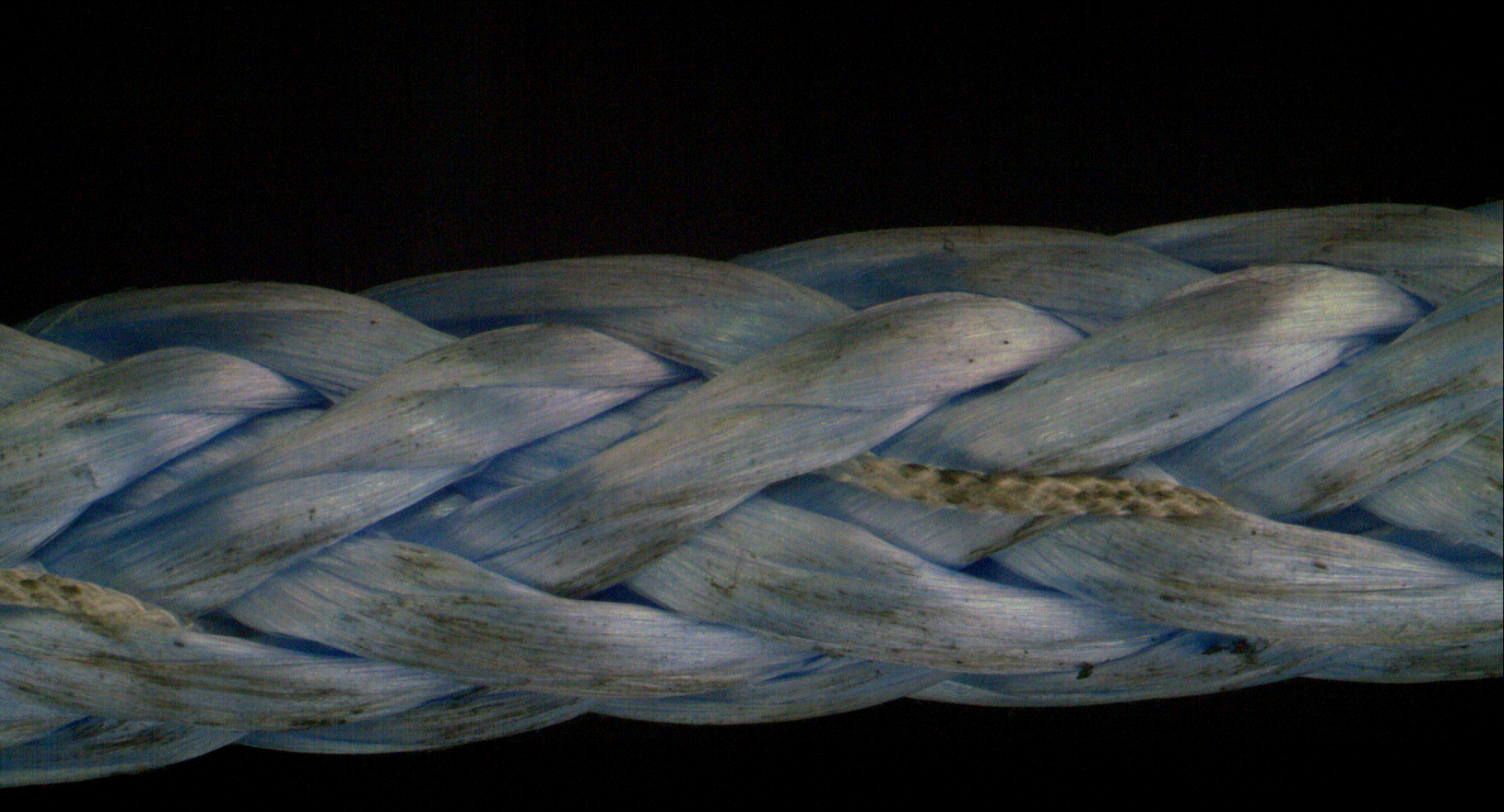}
         \caption{Core out}
         \label{fig:Defects13}
     \end{subfigure}
      \begin{subfigure}[b]{0.32\textwidth}
         \centering
         \includegraphics[width=\textwidth]{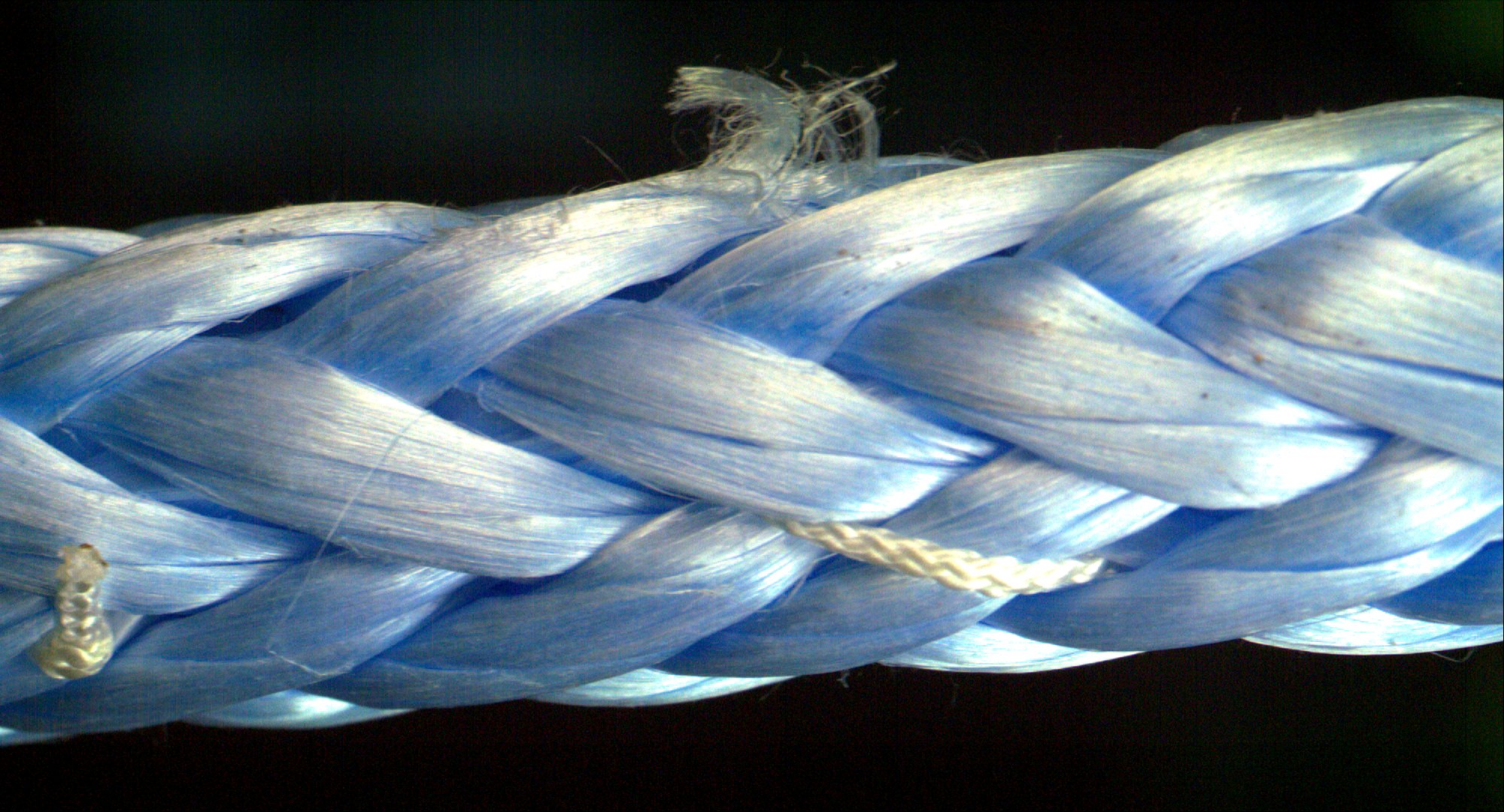}
         \caption{Coreout with Cut Strands}
         \label{fig:Defects14}
     \end{subfigure}
     \begin{subfigure}[b]{0.32\textwidth}
         \centering
         \includegraphics[width=\textwidth]{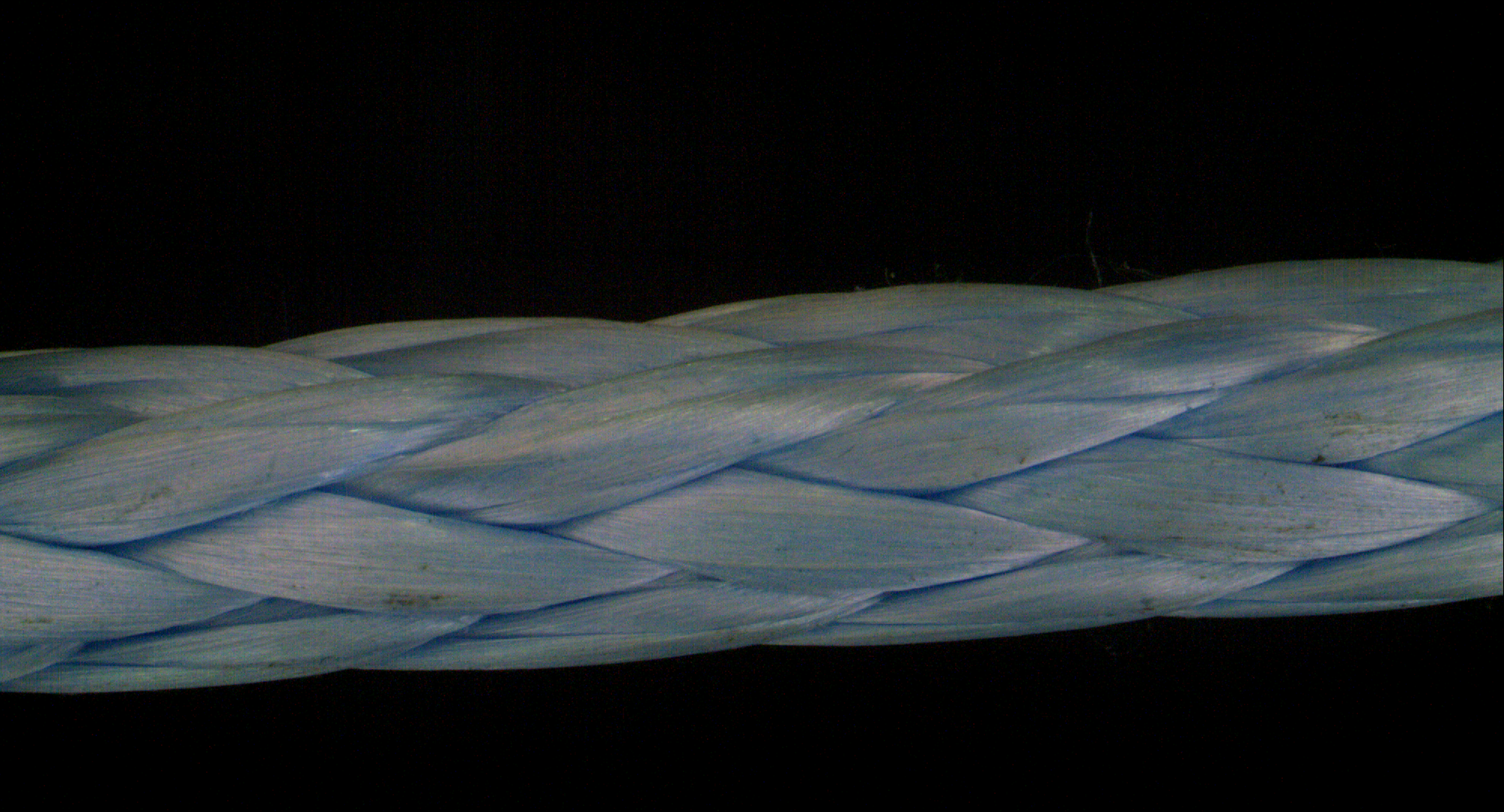}
         \caption{Normal}
         \label{fig:Defects15}
     \end{subfigure}
        \caption{Different defects scenarios considered in the SFRs dataset.}
        \label{fig:Defects}
\end{figure}

\section{Objective}
Synthetic fibre ropes, such as HMPE (high modulus polyethylene) and UHMWPE (ultra-high-molecular-weight polyethylene), offer advantageous properties as alternatives to steel wire ropes (SWRs) in underwater equipment and heavy-load handling applications \cite{rani2023defect, mckenna2004handbook, liftra}. The high resistance to frictional wear, high tensile strength, lightweight nature, and flexibility make them a desirable choice. Effective CM of these SFRs is crucial for diagnosing and preventing system malfunctions, as well as forecasting reliability and determining the RUL of the ropes. To support the development of robust defect detection algorithms and methodologies, this imagery dataset has been generated, encompassing various possible defect scenarios in SFRs. The dataset serves as a benchmark for evaluating the performance of different defect detection techniques. This dataset aims to foster collaboration and knowledge sharing among industry experts, researchers, and stakeholders involved in SFR's applications. Also, the availability of this dataset can encourage the adoption of standardized testing protocols and best practices for SFR's inspection across industries. By utilizing the dataset's annotations, researchers can explore a wide range of deep learning methodologies, including classification and defect detection tasks. The dataset promotes collaboration, standardization, and innovation in the inspection and condition monitoring of SFRs, facilitating the adoption of advanced techniques and best practices across industries.

\section{Data Description}
The dataset has a total of 6,942 labeled images having a dimension of 2000 x 1080 pixels in PNG format indicating the defective and normal condition of the SFRs. The defective ropes are compiled based on the defect types and their severity (high, medium and low). In the repository, data has been compiled into six separate folders; one folder is for non-defective ropes in zip format while the other five folders are in zip format for plackings, cut strands, chafing, compression, and core out defects respectively. Each defective folder is further divided based on their defect type or their severity level if any named as high, medium and low respectively. \textbf{Placking} folder contains all images of the defect class named as \emph{placking high, placking medium and placking low}. The same criteria have been performed for the \textbf{cut strands} and \textbf{chafing} defects. \textbf{Compression} folder contains images of the ropes with irregular rope diameters with three separate classes named \emph{compression, compression with chafing and compression with cut strands}. The \textbf{core out} folder contains images of the SFRs with the centre core on the rope surface with two classes \emph{core out and core out with cut strands}. Lastly, the \textbf{Normal} folder contains images of the class \emph{normal}. A detailed description of defects and defect class in the SFRs dataset has been mentioned in Table \ref{table:Defect_class}. 

\begin{figure}[ht]
\centering
\includegraphics[width=\textwidth]{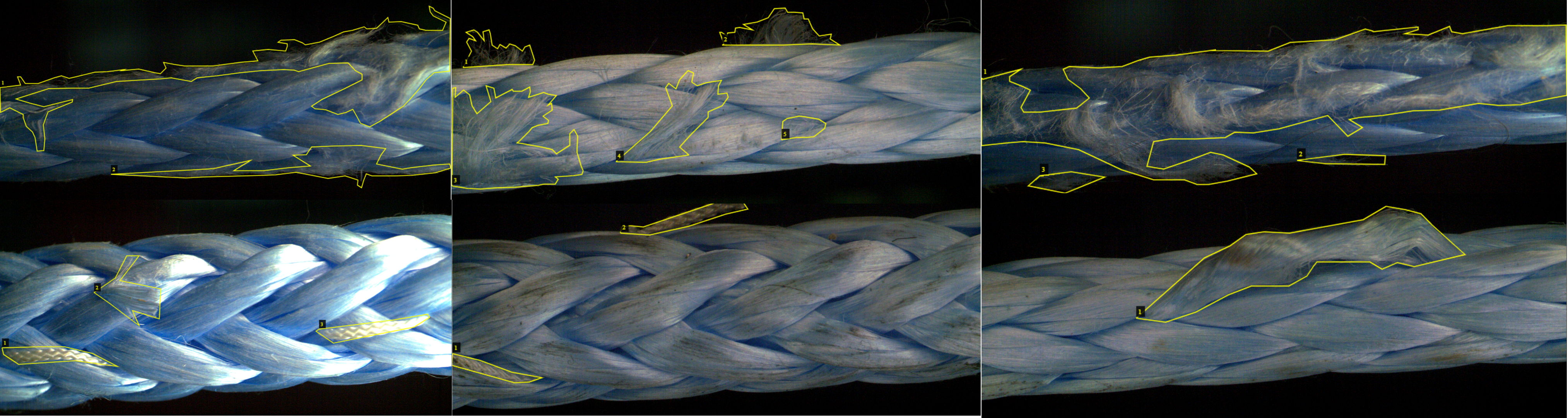}
\caption{Sample of the annotated SFRs dataset.}
\label{fig:Annotated}
\end{figure}

\section{Experimental Design, Materials and Methods}
The experimental setup involved a motor, four LED lights,  three sheaves (two rotation pulley blocks, two wire guide wheels and one sheave for holding weight), and ten SFRs each of length 8 cm subjected to a weight of 50 kg while rotating through a pulley system. To ensure proper lighting for image capture, four Aputure AL-MC RGBWW LED lights were used, each providing an illumination of 1000 lux. These lights were strategically positioned at an angle of approximately 45 $^{\circ}$ to ensure uniform illumination during image capture.
During the data collection process, the SFRs were rotated on the sheaves, which were supported by rotation pulleys and wheels designed to guide the rope used for lifting. This rotational movement aimed to simulate real-world scenarios where ropes are subjected to rotational forces and movement during lifting operations. By replicating these conditions, the data collection process aimed to capture a more accurate representation of the rope's behaviour and the potential defects that may arise during lifting and operational activities. The experimental setup used for collecting the dataset has been illustrated in Figure ~\ref{fig:Setup}.

\subsection{Rope Description}
Dyneema is a multi-filament fibre made through a gel-spinning process, derived from materials like HMPE or UHMWPE. It boasts a range of remarkable qualities including impressive strength, lightweight properties, minimal elongation at break, and resilience against various chemicals and harsh environmental conditions. These outstanding mechanical properties, coupled with its low density, give Dyneema an exceptional performance-to-weight ratio. This makes it a valuable resource for both researchers and industry professionals. It allows them to closely monitor and assess the conditions of fibre ropes, allowing the possibility of substituting traditional SWRs with Dyneema-based alternatives. The SFRs used in the experiment possess the following characteristics:
\begin{itemize}
\item  Fibers: Dyneema SK 75/78
\item  Nominal Diameter: 8 mm
\item  Construction: 12 strands / 12 braided rope
\item  Torsional neutral: The rope is designed in a way to resists twisting or torsional forces.
\item  Pitch/stitch length: Approximately 11mm
\item  Braiding period: Approximately 66mm
\end{itemize}

\subsection{Defects Description}
The experiment has been performed on ten SFRs where artificial defects have been introduced by an expert roper from Dynamica Aps, Denmark \cite{dynamica} to create a diverse range of defect scenarios to facilitate defect detection and evaluation algorithms. Out of the ten ropes, nine ropes are considered defective, while one rope served as a non-defective or normal reference. The defects were distributed among the nine defective ropes, following a pattern. These defects included placking defects with varying severity levels (high to medium) in three ropes, cut strands in another three ropes, and chafing defects in the remaining three ropes. Additionally, each of the nine defective ropes also featured compression and core-out defects. By introducing these specific defects in the experiment, the goal was to provide a realistic representation of the types and severity levels of defects that can occur in SFRs during their operational lifespan. This approach allowed for the development and evaluation of defect detection algorithms in real-world scenarios. Furthermore, the ISO standard 9554:2019, titled "Fibre rope – General specifications," was referred for information about potential damages on SFRs \cite{ISO9554}. This ISO standard served as a valuable reference, providing insights into the types of defects that can manifest in these ropes, aligning with the experiment's objectives. Figure ~\ref{fig:Defects} depicts the artificially introduced defects in the SFRs offering a clear illustration of the various defect types encountered during the experiment.

\subsection{Data Collection}
The image dataset has been collected using a Basler acA2000 camera with a Basler C11-5020-12M-P Premium 12-megapixel lens. To read the images from the camera, an NVIDIA Jetson Nano P3450 was used as the processing (for reading and analyzing the captured data) platform. A total of 6,942 images having a resolution of 2000 x 1080 pixels have been collected to apply the defect detection algorithms. The collected images were then organized and renamed according to their respective classes before being uploaded to an open-access repository. The dataset includes various classes such as normal, compression, core out, placking (high, medium, low), cut strands (high, medium, low), and chafing (high, medium, low) respectively. The labeling of each image was performed using the VGG image annotator (VIA) \cite{dutta2016vgg, dutta2019vgg}, an open-source tool. Figure ~\ref{fig:Annotated} depicts the annotated images of the SFRs obtained from the VIA. The final dataset has been deposited in the Mendeley repository, ensuring its availability for further research and development. The dataset provides high-quality images of SFRs suitable for object detection and segmentation tasks \cite{ridge2001effect}. Researchers can utilize this dataset to develop and enhance algorithms and methodologies in the field of SFR analysis, enabling more accurate and efficient detection and segmentation of defects \cite{debeleac2020experimental}.

\section*{Ethics Statement}
This research does not involve experiments, observations, or data collection related to human or animal subjects. 

\section*{Declaration of competing interest}
The authors declare that they have no known competing financial interests or personal relationships that could have appeared to influence the work reported in this paper.

\section*{Data Availability}
\href{https://data.mendeley.com/datasets/by9wy6fxsr} {Imagery Dataset for Condition Monitoring of Synthetic Fibre Ropes} (Mendeley Data). 

\section*{Acknowledgement}
This research was supported by Aalborg University, Liftra ApS (Liftra), and Dynamica Ropes ApS (Dynamica) in Denmark under the EUDP program through project grant number 64021-2048.

\bibliographystyle{unsrtnat}
\bibliography{references}  
\end{document}